\documentclass{article}
\usepackage{arxiv}

\usepackage[utf8]{inputenc} 
\usepackage[T1]{fontenc}    
\usepackage{hyperref}       
\usepackage{url}            
\usepackage{booktabs}       
\usepackage{amsfonts}       
\usepackage{nicefrac}       
\usepackage{microtype}      
\usepackage{lipsum}

\usepackage{cite}
\usepackage{amsmath,amssymb}
\usepackage[linesnumbered,ruled]{algorithm2e}
\usepackage{graphicx}
\usepackage[makeroom]{cancel}
\usepackage{mathtools}
\usepackage{subcaption}
\usepackage{enumitem}
\usepackage{balance}
\usepackage{tabularx}
\usepackage{color, colortbl}
\usepackage{algpseudocode}

\definecolor{Gray}{gray}{0.9}

\usepackage{amsmath}

\title{Generative Neural Network based  Spectrum Sharing using  Linear Sum Assignment Problems}
\author{
  Ahmed B.Zaky \\
  Big data Institute\\
  School of Computer Science\\
  Shenzhen University,China\\
  \texttt{ahmed.zaky@szu.edu.cn}\\
  Benha University, Egypt\\
   \texttt{ahmed.zaky@feng.bu.edu.eg}
   \And
 Joshua Zhexue Huang \\
 Big data Institute\\
  School of Computer Science\\
  Shenzhen University,China\\
  \texttt{zx.huang@szu.edu.cn}\\
    \And
 Kaishun Wu \\
  Guangdong Laboratory of Artificial Intelligence \\ and Digital Economy (SZ)\\
  Shenzhen University,China\\
  \texttt{wu@szu.edu.cn}\\
  PCL Research Center of Networks and Communications\\, Peng Cheng Laboratory, China.
    \And
    Basem M.ElHalawany \\
   Guangdong Laboratory of Artificial Intelligence \\ and Digital Economy (SZ)\\
  Shenzhen University,China\\
  \texttt{basem.mamdoh@szu.edu.cn}\\
  Benha University, Egypt\\
   \texttt{basem.mamdoh@feng.bu.edu.eg}
}

\begin{document}
\maketitle

\begin{abstract}
Spectrum management and resource allocation (RA) problems are challenging and critical in a vast number of research areas such as wireless communications and computer networks. The traditional approaches for solving such problems usually consume time and memory, especially for large-size problems. Recently different machine learning approaches have been considered as potential promising techniques for combinatorial optimization problems, especially the generative model of the deep neural networks. In this work, we propose a resource allocation deep autoencoder network, as one of the promising generative models, for enabling spectrum sharing {\color{black} in underlay device-to-device (D2D) communication} by solving linear sum assignment problems (LSAPs). Specifically, we investigate the performance of three different architectures for the conditional variational autoencoders (CVAE). The three proposed architecture are the convolutional neural network (CVAE-CNN) autoencoder, the feed-forward neural network (CVAE-FNN) autoencoder, and the hybrid (H-CVAE) autoencoder.
The simulation results show that the proposed approach could be used as a replacement of the conventional RA techniques, such as the Hungarian algorithm, due to its ability to find solutions of LASPs of different sizes with high accuracy and very fast execution time. Moreover, the simulation results reveal that the accuracy of the proposed hybrid autoencoder architecture outperforms the other proposed architectures and the state-of-the-art DNN techniques.
\end{abstract}


%

\section{Introduction}
Radio resource management (RRM) in wireless communications aims to improve the system performance. RRM plays a vital role in many communication systems including machine-to-machine (M2M) communications, device-to-device (D2D) communication \cite{M2M}, antenna selection (AS) in multiple-input multiple-output (MIMO) systems \cite{Antenna_selection,Antenna_selection2}, sub-carriers allocation in orthogonal frequency-division multiple access (OFDMA) based communications \cite{relay_OFDMA}, relay selection in cooperative communication networks \cite{relay_OFDMA}, and spectrum sharing in underlay device-to-device communication \cite{2D2DMealaway}. Many RRM techniques in wireless communications can be cast in the form of an assignment problem at which a number of items are assigned/allocated to a number of workers in an optimal way to maximize the system performance (i.e., minimize or maximize certain utility functions such as sum rate, outage probability, energy-efficiency, and number of admitted users). The process of acquiring an optimal solution is sometimes complex and time-consuming, especially in NP-hard mixed integer optimization problems with large number dimensions. Seeking a fast and low-complexity sub-optimal solution for the assignment problems is an attractive and hot topic in different research areas including wireless communication, computer networks, and mobile computing.

On the other hand, deep learning (DL) has been widely used in different layers in various wireless communication systems. DL techniques could provide intelligent functions that adaptively exploit the wireless resources, optimize the network operation, and guarantee the QoS needs in real-time \cite{ML_Survey1,ML_Survey2}. Recently, DL approaches have been used as a mean of solving several mathematical optimization problems \cite{Pilot_Assignemnt,Q_LTE_U}. In \cite{Pilot_Assignemnt},  the authors proposed a deep learning-based pilot assignment scheme (DL-PAS) for a massive multiple-input multiple-output (massive MIMO) systems for multiple users. In \cite{Q_LTE_U}, the authors investigated a distributed Q-learning mechanism that exploits prior experience for selecting the appropriate channel to use for downlink traffic offloading in the unlicensed 5 GHz band of the Long Term Evolution (LTE-U). The author in \cite{paking_Assig} proposed a deep learning approach for optimizing a 1D variable sized bin packing assignment problem. In \cite{BW_latency},  artificial neural network (ANN) is applied in learning uplink transmission latency, thereby in achieving a flexible bandwidth allocation decisions that reduce latency. Almost all introduced communication problems achieve prominent results using machine learning techniques, which lead us to introduce this work for efficient combinatorial optimization problems using deep learning approach.

{\color{black}The process of RRM even becomes more complicated in heterogeneous network architectures with massive number of nodes as in D2D-enabled networks. In D2D communication, direct transmissions between neighboring nodes are allowed without passing messages through the central base station (BS), which enhances the achievable rates due to proximity services, and improves the energy efficiency by reducing the transmission power \cite{2D2DMealaway,D2D_Hung,BasemDVT}. Recently, two modes of operations have been presented for D2D communication, namely the overlay and underlay mode. In overlay mode, dedicated resources are allocated for D2D transmitters  to avoid interference with traditional cellular users (CUs). On the other hand, direct transmissions between D2D pairs are allowed in underlay D2D mode by sharing the resources (i.e., sub-channels or resource blocks) of cellular users. Each D2D transmitter is allowed to share one or more sub-channel of the available CU, under the control of the central BS, as long as the quality-of-services (QoS) of both nodes are satisfied. If one D2D transmitter is allowed to share the resources of one CU, one-to-one matching problem could be formulated in the form of linear sum assignment problem (LSAP) as in \cite{D2D_Hung}. }

\subsection{Linear Sum Assignment Problems (LSAPs)}
In general, there are different types of assignment problems which can be classified based on the nature of the objective function to be optimized (linear, quadratic,..etc.) and the number of sets to be matched ( 2: two-dimensional assignment problems, 3 or more: multi-dimensional assignment problems). In this work, we investigate a deep learning-based approach for the LSAP. The LSAP is a special case of 2D assignment problems with a linear objective function, which can be considered as a classical combinatorial optimization problem and can be widely found in many wireless communication systems as a stand-alone problem or as a part of a 3D assignment problem. The 3D assignment problems can be solved by decomposing the complex 3D mapping problem into three 2-dimensional sub ones and solve them iteratively \cite{3DAP}. In LSAP, we have 2 sets of items (n-jobs and n-workers) and the assignment process can be viewed as a bijective mapping $\phi$ between the two finite sets where each worker is assigned to one job at most. {\color{black}Notice that it is not necessary for the two sets to be of equal size. If the number of elements is different, several dummy elements are usually used to obtain a square cost matrix. As an example, if the number of workers ($n$) is larger than the number of jobs ($m$),  a group of ($n-m$) dummy jobs are added to have a square matrix which means that the workers assigned to those dummy jobs will be idle at the end of the assignment}. The bijective mapping process  can be written as a permutation as follows
\begin{equation}\label{permute}
\begin{pmatrix}
1 & 2 &...& n \\
\phi\left(1\right) & \phi\left(2\right) &...& \phi\left(n\right)
\end{pmatrix}
\end{equation}
where the permutation means that the index $n$ maps to $\phi\left(n\right)$. Permutation can also be written for example as $\phi = \left(3,2,5,...,1\right)$ which corresponds to the set $I = \{\,i\,|\,1,2,....,n\}$. Notice that each permutation $\phi$ corresponds in a unique representation of an $n\times n$ assignment matrix $X_\phi =\{x_{ij}\}$ where $x_{ij} = 1$ if $j = \phi(i)$ and $0$ otherwise \cite{Burkard_book}. The LSAP optimization problem is usually described using the permutation matrix as
  \begin{align}\label{LSAP_EQ}
   \nonumber \underset{\{x_{ij}\}}{\mbox{min}}  \quad &\quad\sum_{i=1}^{n}  \sum_{j=1}^{n} x_{ij} \, c_{ij}, \\
   \nonumber\mbox{s.t.}\,\,  &C_1: \sum_{i=1}^{n} x_{ij} = 1, \quad \forall j=1,....,n, \\
    \nonumber &C_2:  \sum_{j=1}^{n} x_{ij} = 1, \quad \forall i=1,....,n, \\
    &C_3: x_{ij} \in \{0,1\}, \quad\forall i,j=1,....,n.
  \end{align}
where $c_{ij}$ is the cost of assigning the $i^{th}$ and the $j^{th}$ elements given that $C = \{c_{ij}\}$ is the $n\times n$ cost matrix. The objective is to seek for an assignment with the minimum total cost which can be written as $\sum_{i=1}^{n}c_{i\phi(i)}$. The group of constraint $C_1$ means that each worker $j$ is assigned to only one job at most, $C_2$ means that each job $j$ is assigned to only one worker at most, and $C_3$ is the binary selection index between $i$ and $j$.

Many algorithms have been proposed for solving LSAPs since 1940s. To the best of the authors knowledge, Easterfield \cite{Easterfield_Alg} proposed the first non-polynomial $\mathcal{O}(2^n n^2)$ time approach for LSAP in 1946. In 1950s, Kuhn has proposed the original form of the famous Hungarian algorithm which solves the problem in $\mathcal{O}(n^4)$ time by combining the graph theory and the duality of linear programming \cite{Kun_Hung}. A variations of the Hungarian  algorithm was presented by Munkres \cite{Munkres_Hung} and followed by Edmonds and Karp \cite{Karp_Hung} which reduced the time complexity for LSAP to $\mathcal{O}(n^3)$.

On the other hand, different algorithms have been developed to achieve near-optimal solutions for the LSAPs. First, a couple of heuristic algorithms have been proposed to achieve a sub-optimal solution under a time constraint such as the greedy randomized adaptive algorithm \cite{GRASP} and the deep greedy switching algorithm \cite{subopt_deep_greedy}. Second, the authors in \cite{BNNLSAP} have proposed a DL-based approach to exploit the recent breakthroughs of deep neural network (DNN) models. A complex DNN have been introduced to find a sub-optimal solution for the LSAP by proposing a decomposition approach for the original problem where each sub-problem is turned into a classification task. In each sub-problem, they adopt a supervised learning scheme to nonlinearly maps the input/output relation of the corresponding row of the output $n \times n$ assignment matrix, $X_\phi$, individually. This decomposition approach is followed by a low-complexity heuristic algorithm (a collision avoidance rule) to combine the outputs of each sub-assignment problem together and try to find a global solution. They investigated two different DNN architectures: the feed-forward neural network (FNN) and the convolutional neural network (CNN).

\begin{itemize}
\item {Motivations and Contributions}
\end{itemize}

Major improvements in the wireless communication systems efficiency require making the best use and distribution of the available resources. However, RRM algorithms may have a high price in terms of the complexity and the time constraints. Motivated by this fact, we investigate a deep neural network approach for solving the LSAP as one of the widely used assignment problems to seek a sub-optimal solution with high accuracy and low computation time. In this paper, we propose a generative model based on deep autoencoder to solve assignment problems in high accuracy. The autoencoder is artificial neural network developed to find out data representation in unsupervised learning approach which has been used for a dimension reduction of images, audio, and video signals. Deep autoencoders have also been used for a number of tasks in DNN-based wireless communication algorithms \cite{Autoencoder}. There exists different types of  generative approaches that utilize the autoencoder such as the adversarial autoencoder which uses the generative adversarial networks (GAN) \cite{Goodfellow2014Generative}, the variational autoencoder (VAE) \cite{Kingma2013Auto}, the importance weighted autoencoders \cite{Burda2015Importance}, and the conditional variational autoencoder (CVAE) \cite{Kingma2014Semi}. The main contributions in this work can be listed as follow:
\begin{itemize}
    \item Employ the conditional variational autoencoder (CVAE) for solving the two-dimensional LSAPs.
	\item Propose three neural network architectures for the proposed deep CVAE autoencoder, namely the convolutional neural network (CNN-CVAE), the feed-forward neural network (FFN-CVAE), and the hybrid (H-CVAE) autoencoder.
    \item Investigate the performance of the proposed techniques in terms of the accuracy and time-consumption compared to the conventional well-known Hungarian algorithm and the state-of-the-art DNN model in \cite{BNNLSAP}.
\end{itemize}
The proposed model has different characteristics from that in literature as follows:
\begin{itemize}
	\item In \cite{BNNLSAP}, the investigated a classification-based model, where the output assignment matrix is decomposed into n classification sub-problems which is followed by a collision avoidance step, while our proposed model uses different deep autoencoder architecture to solve the LSAP.
    \item The classification based model has a major scalability problem based on the simulation results in \cite{BNNLSAP} where the accuracy drops for LSAPs of size $n > 4$ due to the divergence from the optimal solution, while the learning methodology of the proposed model using the autoencoders guarantees extracting important adequate features from the original data in order to solve the LSAP.
\end{itemize}

The rest of the paper is organized as follows: In section II, we introduce a case study for LSAP in the area of wireless communication by explaining the system model, parameters, generation of the cost matrix, and training dataset. In section III, the different types and architectures of the autoencoders are presented. Section IV presents the proposed CVAE model architecture, the implementation details and the training methodology of the proposed autoencoders. In section V, we provide the experimental results. Finally, the paper is concluded, and the future work is listed in section VI.
\begin{figure}[!t]
	\centering
	\includegraphics[height=9cm, width=0.6\columnwidth]{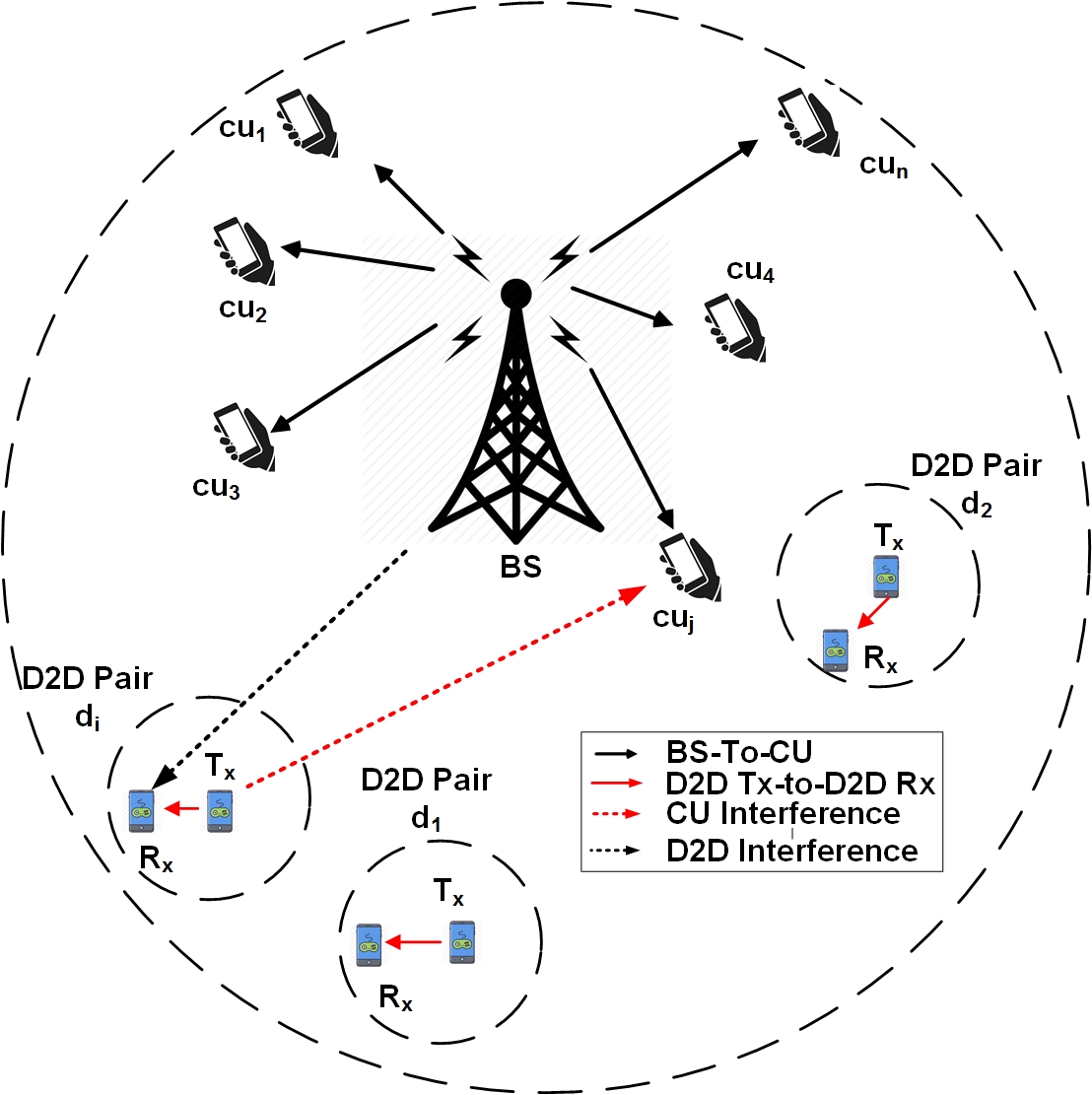}
	\caption{ A sample application of LSAPs in wireless communication systems where a group of D2D pairs coexists with a group of  cellular users in a cellular network.}
	\label{D2D_model}
\end{figure}
\section{Case Study for LSAP: Underlay Device-to-Device Communication}
Recently, the fifth generation (5G) of wireless systems has introduced many promising applications, where multiple technologies coexist together for improving the performance. In the following subsections, we explain in detail a case study on the resource allocation of an underlay device-to-device (D2D) communication scenario similar to the model in \cite{D2D_Hung}. We summarize the optimal resource allocation algorithm introduced in \cite{D2D_Hung}, emphasis on the LSAP nature of the problem, generation of the cost (weight) matrices taking into consideration the system parameters, and the optimal solution using the conventional Hungarian algorithm. Finally, we describe our  datset, which is used to train the proposed DNN models to attain a fast and near optimal solution.

\subsection{Resource Allocation for Underlay D2D Communication}
 {\color{black}In the following, we briefly describe the underlay D2D system model investigated in \cite{D2D_Hung} for the completeness of the work.

 We consider a single cell scenario as shown in Fig. \ref{D2D_model}, which consists of central base station (BS), a group of $m$ D2D pairs (i.e., each pair includes one D2D transmitter and one receiver node), and a group of $n$ traditional downlink cellular users, where $n>m$. The set of cellular users is represented
 as $C = \{cu_j\}$ for $j\in\{1,...,n\}$, whereas the set of D2D pairs is
 represented as $D = \{d_i\}$  for $i\in\{1,...,m\}$. We assume an underlay D2D  communication mode, where the D2D transmitters are allowed to share the downlink resources of the cellular user. This means that D2D receivers experience interference from the downlink transmission from the BS to CUs, while each CU experiences interference from the D2D transmitter that reuses the same resource block and no interference from other CUs. Since sharing the resource block of a CU with multiple D2D transmitters incurs higher interference and degradation of the achievable CU rate, we assume that each resource block is shared by one D2D transmitter only. Additionally, allowing the D2D pairs to reuse multiple CUs leads to complex models for the power distribution and complex processing; hence each D2D pair is allowed to share resource block of one CU at maximum.}

We consider a Rayleigh fading path loss model, at which the channel gain between node $x$ and node $y$ is given as $G^{x,y} = 10^{-PL^{x,y}/10}$, where the path loss ($PL$) is given in dB as follows
\begin{eqnarray}
PL = 36.7 \log_{10}(d_{x,y}) + 22.7 + 26 \log_{10}(f_c),
\end{eqnarray}
 where $d_{x,y}$ denotes the distance in meters between $x$ and $y$, $f_c$ is the frequency in GHz. The signal-to-interfernce-plus-noise ratios (SINRs) at the $j^{th}$ cellular user and the $i^{th}$ D2D receiver are given respectively as follows
 \begin{eqnarray}
\displaystyle \gamma_j &= \frac{P_B\, G^{B,j}}{N + \sum_{i} x_{i,j}\, P_i\,G^{i,j}}\\
\displaystyle \gamma_i &= \frac{P_i\, G^{i}}{N +  x_{i,j}\, P_B\,G^{B,i}}
 \end{eqnarray}
 where $P_B$ denotes the transmission power of BS to each CU, $P_i$ is the transmission power of the $i^{th}$ D2D transmitter, $N$ is the additive white Gaussian noise (AWGN) power spectral density at the receivers, $G^{B,j}$ is the channel gain between the BS and $cu_j$, $G^{i,j}$ is the channel gain between the $i^{th}$ D2D transmitter and $cu_j$, $G^i$ is the channel gain between transmitter and receiver of the $i^{th}$ D2D pair,  $G^{B,i}$ is the channel gain between the BS and the $i^{th}$ D2D pair receiver, and $x_{i,j}$ is a binary assignment variable, which equals one if the $i^{th}$ D2D pair is reusing $cu_j$'s resource block and zero otherwise. The achievable rate of $cu_j$ can be expressed using Shannon's capacity formula as $R_j = B\,\log_2(1+\gamma_j)$ if its resource block is shared with a D2D pair and $R_j^o = B\,\log_2(1+\frac{P_B\, G^{B,j}}{N})$ otherwise, where $B$ is the channel bandwidth. The achievable rate of the $i^{th}$ D2D receiver can be similarly expressed as $R_i = B\,\log_2(1+\gamma_i)$.

\begin{figure}[!t]
	\centering
	\includegraphics[height=5cm, width=0.5\columnwidth]{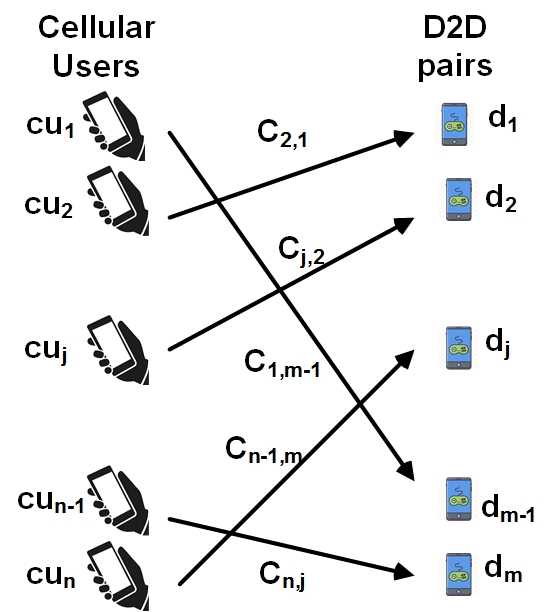}
	\caption{One-to-one matching between cellular users and D2D pairs.}
	\label{one2one}
\end{figure}

 The aim of the resource allocation is to maximize the total system sum rate of all CUs and D2D pairs under QoS constraints on the achievable SINRs for both CUs and D2D pairs, which can be formulated as follows
 \begin{subequations}
 	\begin{align}
	&\underset{\{x_{i,j}\}}{\max}\qquad \sum_{j=1}^{n}\sum_{i=1}^{m} (R_j + x_{i,j} \,R_i)\label{D2d_Ob}\\
	\nonumber &\mbox{subject to}\\
	&\qquad \gamma_j \ge \gamma_j^{th},~\quad\qquad \forall j\in\{1,...,n\}\label{D2d_a}\\
	&\qquad \gamma_i \ge \gamma_i^{th}, ~\quad\qquad\forall i\in\{1,...,m\}\label{D2d_b}\\
	&\qquad R_j + x_{i,j}\, R_i \ge R_j^o, \forall j\in\{1,...,n\}, i\in\{1,...,m\}\label{D2d_d}\\
	&\qquad \sum_{i} x_{i,j} \le 1,\qquad \forall j\in\{1,...,n\}\label{D2d_c}\\
	&\qquad \sum_{j} x_{i,j} \le 1,\qquad \forall i\in\{1,...,m\}\label{D2d_e}\\
	&\qquad x_{i,j} \in \{0,1\},\qquad\forall j\in\{1,...,n\}, i\in\{1,...,m\}\label{D2d_f}
		\end{align}
 \end{subequations}
 where $\gamma_i^{th}$ and $\gamma_j^{th}$ are the QoS threshold values for D2D receivers and CU, respectively. The constraints in \eqref{D2d_a} and \eqref{D2d_b} are used to maintain the QoS requirements, while \eqref{D2d_d} is used to allow  resource sharing if rate enhancement is achievable only. In other word, the CU will not share its resource block with a D2D transmitter unless the summation of the achievable rates of both nodes are better than the rate achieved by the CU without sharing. Notice that the problem can be transformed into LSAP, which can be solved by finding the maximum weight matching in a weighted bipartite graph. The following criteria are used for generating the cost matrices between all D2D pairs and all CUs for solving the bipartite matching problem using the Hungarian algorithm:
 \begin{itemize}
 	\item If constraints \eqref{D2d_a}, \eqref{D2d_b}, and \eqref{D2d_d} are satisfied for the $i^{th}$ D2D pair and $cu_j$, the weight of the edge between a CU and D2D pair shown in Fig. \ref{one2one} represents the sum rate of both nodes (also known as profit or cost element), which is expressed by $C_{i,j} = -(R_j + x_{i,j}\,R_i)$. Notice that the negative sign is because the Hungarian algorithm solves a minimization problem while the objective function in \eqref{D2d_Ob} is a maximization.
 	\item If any constraint is not satisfied, ($-R_j^o$) is assigned to the corresponding edge.
 	\item The Hungarian algorithm solves matching problem with a square matrix. Since we have $n>m$, we add $(n - m)$ dummy D2D pairs starting from row $m+1$. The weight of a dummy D2D pair and $cu_j$ is $(-R_j^o)$.
 	\item  At the end of the algorithm, if any CU is matched with any dummy D2D pair, it means no resource sharing for this CU.
 \end{itemize}

The cost matrix is passed to the Hungarian algorithm, which returns an optimal assignment matrix (also known as label matrix). The rows and columns of the assignment matrix are one-hot vectors that contains only one logic-one and the other elements are logic-zeros. {\color{black}Notice that this model is applicable on any resource allocation scenario as long as it can be re-casted into the form of one-to-one matching similar to the Hungarian algorithm. The main task in such cases (i.e., multi-cell or multi-cluster D2D networks \cite{ElHalawany2019}) is to consider the  rule sets needed to generate the 2D square cost matrix according to the interference management algorithm adopted in those network scenarios, which is followed by using the proposed DNN model. }

\subsection{Dataset Description}
In order to train the proposed DNN models to solve the LSAP, we use the optimal solutions (i.e., assignment matrices) generated by the Hungarian algorithm for a large number of realizations of the described underlay D2D communication scenario. The obtained solutions are used as benchmark and training dataset for the deep network. The size of the generated  dataset is $4\times 10^6$ samples with 90\% of the data to be used as a training set and 10\% for the testing set. Matlab is used for generating the dataset assuming the cell radius $1000$ m, maximum distance between the transmitter and receiver of D2D pairs are 15 m, cellular user and D2D transmit powers are 23 dBm, BS transmit power 46 dBm, Noise power  is -174 dBm, and the carrier frequency is 1.7 GHz. the cellular users and the D2D transmitters are uniformly distributed within the cell area, while the D2D receivers are uniformly distributed within a circle of 15 m radius from the D2D transmitters.

Each sample of the dataset consists of two $n \times n$ matrices, where the first one is the cost matrix and the second one is the boolean assignment matrix. The proposed DNN models are to be trained to generate an optimal solution with a reconstruction error criterion. One of the main challenges are the size of the problem, where for n users, the number of available solutions is the factorial of  $n$ ($n!$). As n increases the samples needed to train the network becomes much larger. As an example, for $n = 8$, the number of available solutions is 40,320.

\section{DNN Autoencoder Architectures}
The recent breakthroughs of the deep neural networks in general and specially in the generative models encourages a lot of researchers to exploit it in different applications. The Generative models are machine learning approaches that learn to estimate the joint probability distribution of $P(x,y)$, where $X$ is an observable variable and $Y$ is a target label variable. The Bayes rule can be used to transfer the joint probability distribution $P(x,y)$ into a conditional probability distribution $p(y|x)$ which is the natural distribution for classifying a sample $x$ into a class $y$. Generative models are able to be used for different machine learning tasks such as classification, clustering \cite{ng2002discriminative}, time series analysis, and generating samples that follow $(x,y)$ pairs relation \cite{Goodfellow2014Generative}. There are different types of generative models which include the Gaussian mixture model (GMM), the Hidden Markov model (HMM), the Naive Bayes, the Latent Dirichlet allocation, the restricted Boltzmann machine, and the generative autoencoder models\cite{Pilot_Assignemnt}.

In this work, we employ the generative autoencoder model which can be considered as a leading approach in unsupervised deep learning for the pre-mentioned samples generation task. The traditional autoencoder uses: (1) an encoder to transform the training input data into a lower dimensional latent form, and (2) a decoder to turn this latent representation back to original data by minimizing a reconstruction loss function in an unsupervised approach. The learning methodology of the traditional autoencoder guarantees the extraction of the important adequate features from the original data. There are different generative models of autoencoders that have been introduced in literature with a considerable success for different applications such as:

\begin{itemize}
\item Adversarial autoencoder (AAE): which is based on  generative adversarial networks (GAN) that uses an adversarial training strategy between two deep networks, a generator and a discriminator network, to populate the output distribution.
\item Variational autoencoder (VAE): which infers the output distribution using Variational Inference (VI) approach. The VI method is a Bayesian inference technique where a simpler distribution is proposed to be used, such as the Gaussian distribution, that is easier in evaluation than the output distribution. VI uses the Kullback\-Leibler (KL) divergence to minimize the difference between those two distributions (i.e., the proposed and the output distributions). The VAE generative model could be used for large datasets.
\item Importance weighted autoencoders (IWAE): is an extension to VAE with similar architecture, but the decoder network uses more than one sample to approximate the output distribution. The main objective of IWAE is to model complex distributions where VAE is not applicable.
\item Conditional Variational autoencoder (CVAE): which is another extension to VAE, where a conditional generative process is employed. The latent representation and the data are conditioned to random variables.

\end{itemize}

The pre-mentioned autoencoder types can be constructed using the following two types of layers:
\begin{itemize}
\item Convolutional layers: which is used to extract the basic features of input data.
\item Fully connected layers: which is used to learn the non-linearity between the features and the mapping between the features and the output.
\end{itemize}

\section{Proposed CVAE DNN Architecture}
In this work, inspired by the ability of the autoencoder to build scalable generative models for data of different distributions, such as images, audio or video, we investigate the ability of deep autoencoders to convert the input data, which represents the cost matrix of the LSAP, to the form of the output data which represents the assignment matrix. A DNN autoencoder with many hidden layers could be used to improve the features extraction capabilities and to efficiently fit more data and a variety of classes. The reason of such expected improvement is the increase in the model training parameters with the increase of the number of layers which introduces an efficient reconstruction of the data with minimal reconstruction errors \cite{Hinton2006Reducing}.

Since the number of possible outcomes of the LSAPs is $n!$ cases which is huge for large-size problems, the traditional training techniques need a huge amount of training samples to cover the outcomes. To tackle this problem, we employ a generative model whose main task is to generate new data samples that follows the same distribution of a given training dataset. Using this generative approach, we train networks with different sizes which are able to globalize and generate other cases that is not trained for.

\begin{figure*}[!t]
\includegraphics[width=0.95\textwidth,height=6.5cm]{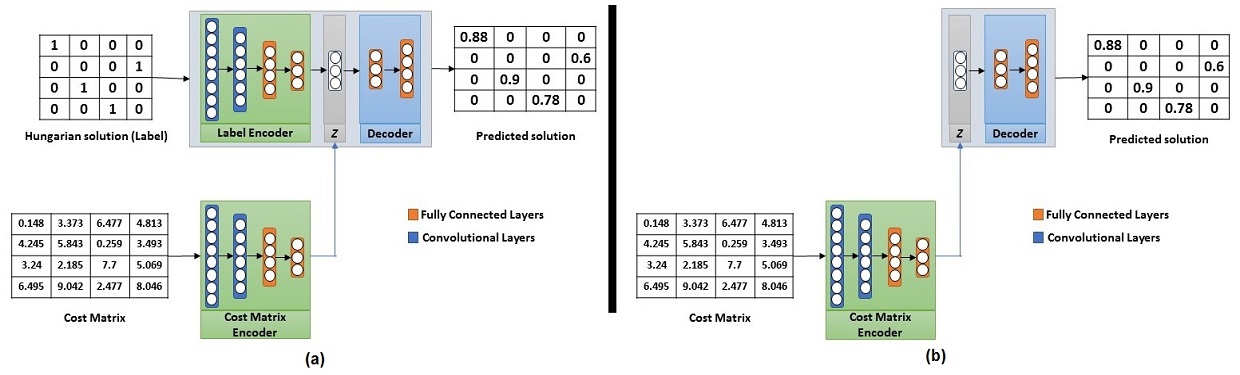}
\caption{Architecture of the conditional variational hybrid autoencoder with the input cost matrix and the output assignment matrix. (a) The training process where we learn the latent representation z for a cost matrix (x) and its optimal solution (y) as a conditional model $y = P(z|x,y,\theta)$. (b) The testing process using the cost matrix only as input.}
\label{Mixed_Arch}
\end{figure*}

Fig.\ref{Mixed_Arch} shows the construction of a conditional variational autoencoder (CVAE) with hybrid structure which is composed of two encoders (The label encoder and the cost matrix encoder) and one decoder. The function of the two encoders is to extract the features of the Hungarian problem solution label and the cost matrix while the function of the decoder is to map the combined features (also known as the latent representation $z$) to the label solution.

The encoders are composed of different layers types, specifically the convolutional and the feed-forward layers while the decoder is composed of fully-connected layers. Notice that in convolutional (CNN) autoencoders and feed-forward (FNN) autoencoders there will be only one type of layers in all sub-component. In this work, we investigate the performance of the three types of autoencoder (i.e., CNN-CVAE, FNN-CVAE, and hybrid-CVAE) for generating the solution of the LSAP and compare it to the state-of-the-art deep network in \cite{BNNLSAP}. Fig.\ref{Mixed_Arch} (a) shows the training phase where the inputs to the hybrid autoencoder are the $n\times n$ cost matrix and the predicted output label corresponding to the input cost matrix while the output is the generated assignment matrix for the LSAP.  Fig.\ref{Mixed_Arch} (b) shows the testing phase of the proposed network at which the label encoder is dropped to use the cost matrix as the only input.

\subsection*{$\boldsymbol{\cdot}$ Model Architecture and Implementation Details}
\begin{figure}[!b]
\center
\includegraphics[width=0.5\textwidth,height=4cm]{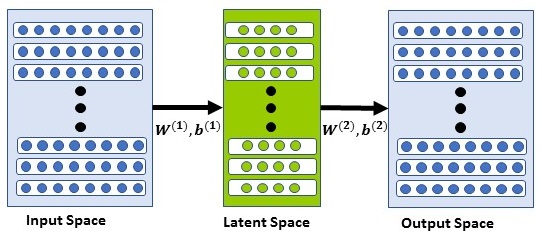} 
\caption{Embedding with a simple three layers autoencoder.}
\label{Embed_Arch}
\end{figure}

Given a dataset constructed as pairs $(X,Y) = \{({x}^{(i)},{y}^{(i)}),...,({x}^{(m)},{y}^{(m)})\}$, where $x^{(i)}$ is the cost matrix of the $i^{th}$ training sample with a corresponding label ${y}^{(i)}$ that represents the optimal solution generated by the Hungarian algorithm. As shown at the simple three-layers autoencoder in Fig. \ref{Embed_Arch}, the objective of the autoencoder is to learn the data embedding representation using the encoder part which have a latent representation $z$. This process is followed by decoding that representation back into the input data. The input layer represents the data in its original high-dimensional form while the latent space layer is the compact embedding representation that contains all the features extracted from the encoder layers.

The autoencoder embedding representation for a given input $x^{(i)}$ is $f(W^{(1)},b^{(1)};x^{(i)}) \in \mathbb{R}$, where $W^{(1)}$ and $b^{(1)}$ are the weight and bias between the input and the latent network layers, respectively and given by:
\begin{equation}
	z = f(W^{(1)},b^{(1)};x^{(i)}) = \sigma(W^{(1)}x^{(i)}+b^{(1)})
\end{equation}
where $\sigma(.)$ is the activation function of the nodes (i.e, ReLu). The network output can be represented as
\begin{equation}
	h(W,b;x^{(i)}) = \sigma(W^{(2)}f(W^{(1)},b^{(1)};x^{(i)})+b^{(2)})
\end{equation}	
where $W^{(2)}$ and $b^{(2)}$ are the weight and bias between the latent and the output layers, respectively. The objective of the loss function $j(W,b;x)$ for a general multilayer autoencoder network is to minimize the output reconstruction errors by using the $L_{2}$ norm. 
\begin{equation}
	j(W,b;x) = \frac{1}{2} \sum_{i=1}^{m} \left \| h(W,b;x^{(i)}) - x^{(i)} \right \| ^ 2
\label{L2_loss}
\end{equation}

The variational type of the autoencoders could learn the low-dimensional embedding $z$ of the cost matrx $x$ by maximizing the lower-bound likelihood in \eqref{KL_Eq} with respect to the group of deep network parameters $\theta$.
\begin{eqnarray}
 \nonumber \log\left(\, P(y|\theta\, ) \,\right)= \mathbb{E}_{z \sim Q} [\log\left(\, P(\,y|z,\theta)\,\right)] 
  - KL[Q(z|y,\theta)||P(z)]
\label{KL_Eq}
\end{eqnarray}
where $\mathbb{E}_{z \sim Q}$ is the expectation of $z$ sampled from $Q$, $P(y|z,\theta)$ is a posterior Gaussian distribution $N(y|f(z;\theta); \sigma_o^2)$ and $P(z)$ is a normal distribution of a zero mean and a unit variance. 

{\color{black}The key idea behind this in variational autoencoder \cite{kingma2013autoBMH} and \cite{kingma2014semiBMH} is to attempt to sample the values of latent $z$ that are likely to produce input $X$, then compute $P(X)$  from those samples. This means that we need a new function $Q(z|X)$ which can take a value of X and give us a distribution over z values that are likely to produce $X$. Hopefully the space of z values that are likely under Q will be much smaller than the space of all z’s that are likely under the prior $P(z)$, which allow us to compute $E_{z \backsim Q}P(X|z)$ relatively easily.}

Therefore, the first term in \eqref{KL_Eq} can be reduced to a decoder network $f(z;\theta)$ with the $L_2$ loss function given in \eqref{L2_loss}. Employing the $L_{2}$ norm as part of the loss function improves the network convergence and helps in avoiding over fitting.

The encoder network $Q(z|x,\theta)$ is trained  with a KL-divergence loss to mimic the distribution $P(z)$. VAE introduces a re-parameterization trick by sampling $z \sim Q$ to support back-propagation  and jointly train the encoder and decoder. The VAE generative model have been successfully used to decode and encode data for various applications such as human faces \cite{Yan2016Attribute2Image}, large scale images CIFAR \cite{Kingma2016Improving}.


In this work, w use the CVAE for solving the LSAP where the model input $x$ is the cost matrix and the output $y$ is the problem solution (optimal solution generated by Hungarian) which is generated from the distribution $P(y|x,z,\theta)$. The CVAE training maximizes the conditional log-likelihood in  where the encoder is additionally conditioned on the cost matrix $x$. The CVAE model can be employed in our case with $ L_{CV}$ in \eqref{Condition_loss_Eq} as the objective loss function.

\begin{eqnarray}\label{Condition_loss_Eq}
 \nonumber L_{CV}(x,y,\theta) = \mathbb{E}_{(z|y,x) \sim Q} \left[\,\log\left( P(y|x,z,\theta )\right)\,\right] 
 - KL\left[\,Q(z|x,y,\theta)\,||\,P(z|x)\,\right]
\end{eqnarray}
 {\color{black} where the prior distribution $P(z|x)$ is assumed to be multivariate Gaussian.} The training and the testing strategies of the proposed CVAE is shown in Fig.\ref{Mixed_Arch}.(a) and (b), respectively where the CVAE embedding $z$ is generated using both cost matrix $x$ and its solution $y$. The decoder network is conditioned on $x$, in addition to $z$. The model testing shown in Fig.\ref{Mixed_Arch}.(b), where the cost matrix is the only input to get the LSAP solution.

Since the problem's solution is a sparse as shown in the table on the right hand side of Fig. \ref{Mixed_Arch}, a rectified linear units (ReLus) are used for the hidden and output layers. The output of each ReLu is 0 if its input is less than 0 and a positive value otherwise.

Regarding the optimization of the objective loss function, the stochastic Gradient Descent (SGD) optimization algorithm is the most popular technique to find the huge number of model parameters. However, there are a lot of challenges exist. As an example, the SGD does not guarantee a good convergence, the selection of a suitable learning rate is difficult, and it needs a lot of trials. Small learning rate leads to a slow convergence, while the large rate can lead the loss function to diverge.
Deep models learned by SGD as in \cite{BNNLSAP} requires more experiments to adjust and find suitable network parameters as the learning rate and the batch size. Additionally, it requires more epochs to convergence as we will show later in experimental results. Moreover, one of the major issues is the use of non-convex loss function that causes the algorithm to be trapped into sub-optimal local minima.

One of the challenges in LSAP is the sparsity of the output where each row and column contain only a single 1 and the remaining values are zeros. As a consequence, SGD is not recommended for learning such problems. On the other hand, the adaptive moment estimation (Adam) \cite{Kingma2014Adam} uses adaptive learning rates for each network parameter which is recommended for the LSAP. Adam is used for the update rule with the default values introduced in \cite{Ruder2016An}.

\section{Experiment Results}
In this section, the simulation results are given to evaluate the performance of the three investigated autoencoder architectures. The three networks are trained for different size of the LASP problem (i.e., n = 4, 8 and 16) where the training data is obtained by using the Hungarian algorithm for the D2D communication scheme in  \cite{D2D_Hung}. {\color{black}The dataset size is  $4 \times 10^6$ and $32\times10^6$ channel realizations (samples) for $n=8$, and 16, respectively. Notice that the number of samples in the dataset is small compared with the number of possible solutions ($n!$) specially for $n = 16$.}
\begin{table}[!b]
\center
\begin{tabular}{|l|l|l|}
\hline
 \textbf{n}     & \multicolumn{1}{c|}{\textbf{Convolutional layers}} & \multicolumn{1}{c|}{ \textbf{Fully connected layers}} \\ \hline
4  & 256 - 128 - 64                                & 512 - 256 - 128 - 64                              \\ \hline
8  & 1024 - 512 - 256 - 128 - 64                       & 4096 - 1024 -512 - 256                           \\ \hline
16 & 4096 - 2048 - 1024 - 512 - 128 - 64                 & 4096-2048-1024-512-256                      \\ \hline
\end{tabular}
\caption{The number of layers and the number of neurons in each layer of the hybrid autoencoder architecture for LSAPs with size n = 4, 8, and 16. }
\label{Net_architecture}
\end{table}

We build nine deep neural networks, i.e. three network types (FNN, CNN and hybrid) for three different sizes ($N = 4,8,$ and $16$). All networks are based on auto-encoding training and consist of one decoder and two encoders. Both encoders have the same architecture for each network. Table I shows the number of layers and the number of neurons in each layer of the hybrid autoencoder in correspondence to the different sizes of the LSAP, where as an example for $N=4$ in the first row, it consists of 3 convolution layers, which are followed by 2 fully-connected layers. The 3 convolution layers have 256, 128 and 64 channels with filter size of $3\times3, 3\times3,$ and $2\times2$, respectively; while the 4 fully-connected layers have 512, 256, 128 and 64 neurons assuming the architecture shown in Fig. \ref{Mixed_Arch}.
	
On the other hand, the other two types (i.e., FNN and CNN) are composed of one type of layers only. The FNN networks encoders use fully connected layers only by dropping convolutional layers from the hybrid design, while the CNN networks encoders use convolutional layers only by dropping fully connected layers from the hybrid design. The decoders for all networks are designed using fully connected layers only with same number of neurons as the decoders but with reverse order. As an example, the hybrid network for n = 4 use four fully-connected layers with 64, 128, 256, 512 neurons.

Table \ref{Net_architecture} shows that as the LSAP problem size grows, more convolutional layers are needed for extracting features efficiently, especially with the millions of training samples. The number of neurons in fully connected layers have to be increased too to store and extract relation between the features and the output. Increasing the size of the LSAP leads to more output combinations (i.e., $n!$) which in turns leads to the requirement of a deeper and wider layers design.
\begin{algorithm}[!t]
\SetAlgoLined
\textbf{Inputs:}  A dataset of $T$ training Samples: ($X_{t}$, $Y_{t}$) where $X_{t}$ is the input cost matrices and $Y_{t}$ is the desired output assignment matrices for $t\in \{1,..., T\}$.
\\ \textbf{Output:} accuracy
\\ $\mbox{accuracy} \leftarrow 0$ ; $\mbox{correct} \leftarrow 0$ ;
\\ $n \leftarrow$ number of rows of $X_t$
\\\textbf{While}($t \leq T$)
\\   $\quad$ $\widehat{Y}_{t} \leftarrow$ observed network output for the sample $t$;
\\   $\quad$ FLAG $\leftarrow 1$
\\   $\quad$ \textbf{For}($\mbox{r} = 1 \mbox{to} \,n$)
\\   $\qquad$ $l(r) \leftarrow$ Index of the only 1 in the $r^{th}$ row of Y[t]
\\ 	 $\qquad$ $\widehat{l}(r) \leftarrow$  Index of highest element in $r^{th}$ row of  $\widehat{Y[t]}$
\\   $\qquad$ \textbf{If} $l(r) \neq \widehat{l}(r)$
\\   $\qquad$ $\quad$ FLAG $\leftarrow 0$
\\   $\qquad$ \textbf{End If}
\\   $\quad$ \textbf{End For}
\\   $\quad$ \textbf{If} FLAG = 1
\\   $\qquad$ $\quad$ correct = correct + 1;
\\   $\quad$ \textbf{End If}
\\\textbf{End While}
\\ accuracy $= \frac{\mbox{correct}}{T}\times 100 \%$
\label{LSAPaccuracy}
 \caption{LSAP accuracy measurement}
\end{algorithm}

In contrast to the deep neural network model in \cite{BNNLSAP}, the binary accuracy measure is not used in our model which can be considered miss-leading. The reason is that the sparsity of the output (As an example, there are 12 zeros and 4 ones in the assignment matrix for an LSAP of size 4) makes the network starts with a binary accuracy measure  of 75\% even if there are no ones in the output at all. We propose an accuracy measure algorithm suitable for the sparse nature of the output assignment matrix. The accuracy of the solution obtained for each  training sample is not obtained for the whole assignment matrix at once. However, we iterate on each row of the assignment matrix and search for the location with the highest probability in this row. Then, we compare that location with the one-hot location in the corresponding row of the training sample. The result is not counted as accurate one unless the locations in all rows are similar to those of the training sample. The details of the accuracy measure algorithm is shown in Algorithm I.

\begin{figure}[!t]
	\centering
	\includegraphics[width=0.6\linewidth,height=5cm]{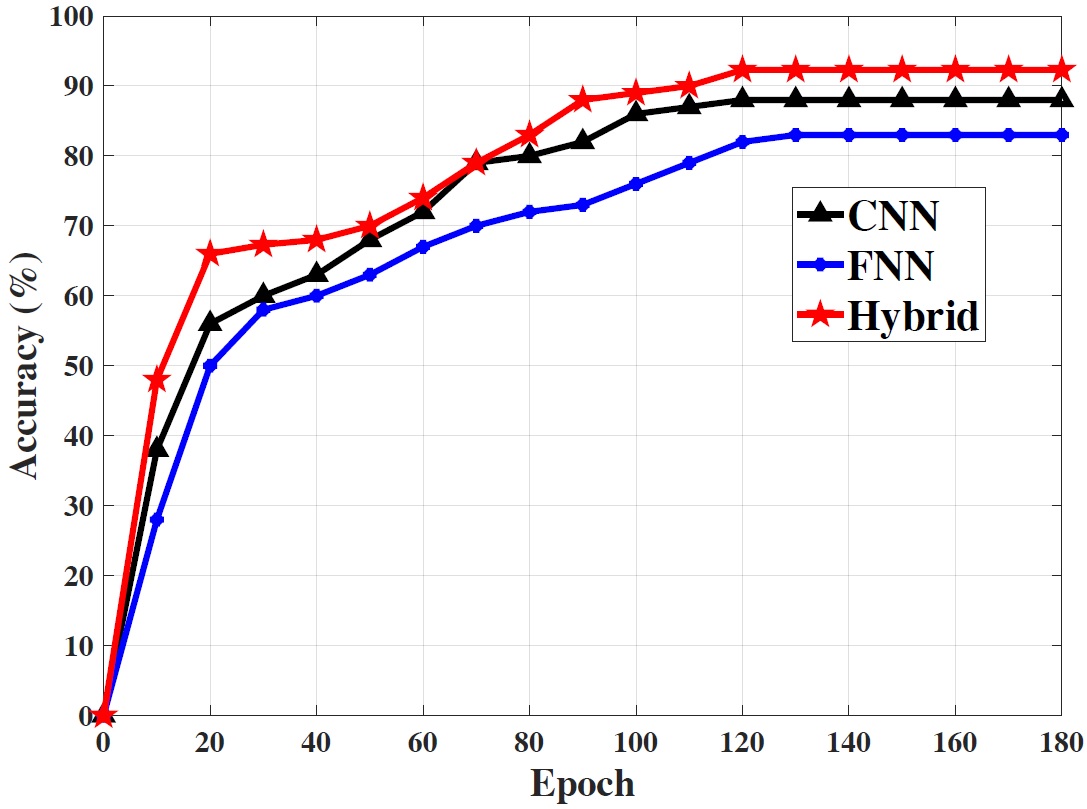}
\caption{The accuracy of the three investigated deep neural network CAVE autoencoders for a LSAP of size n = 16 as a function of the training epoch.}
\label{accuracy}
\end{figure}

\begin{figure}[!b]
	\centering
\includegraphics[width=0.6\linewidth,height=5cm]{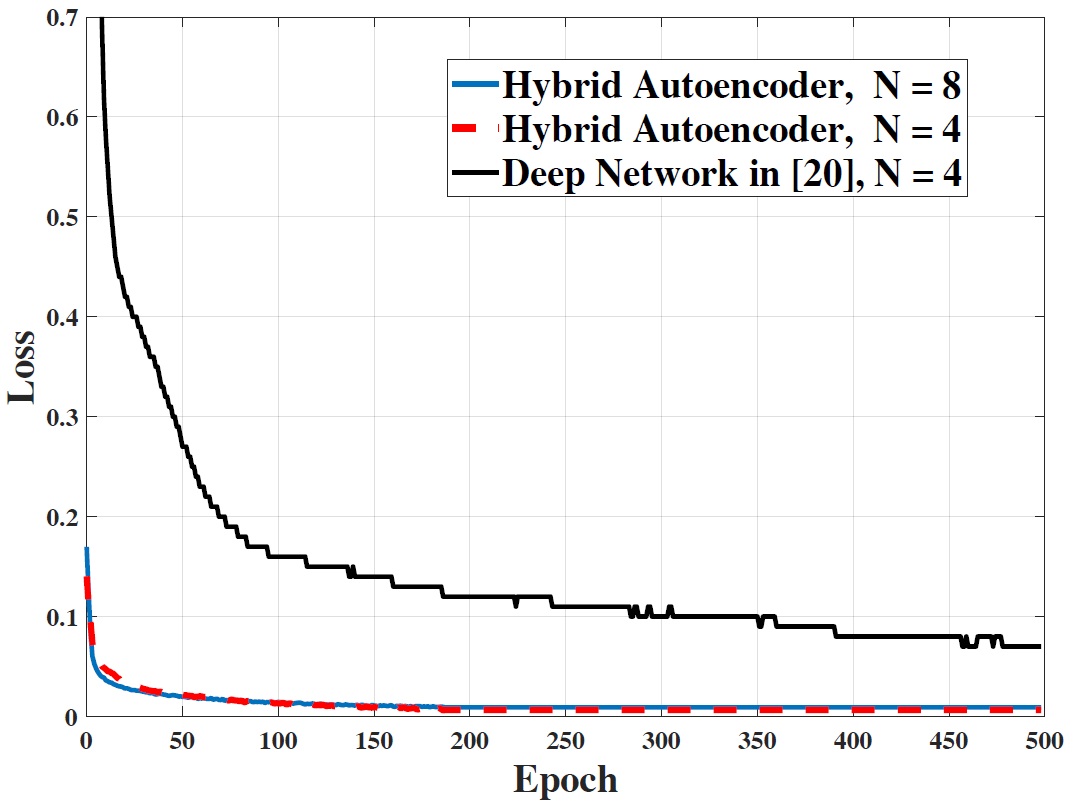}
\caption{The training loss values for the proposed hybrid CAVE autoencoder for n = 4 and 8 compared to the best result of the DNN model in \cite{BNNLSAP} for n = 4.}
\label{loss}
\end{figure}

Fig. \ref{accuracy} shows the convergence of the accuracy of the solutions obtained using the three investigated autoencoders for a LSAP of size 16 as a function of the training epoch. The simulation results show that the three architecture achieve an accuracy of 83, 88.7, and 92.3 \% for the feed-forward, convolutional, and the hybrid deep neural autoencoder architectures, respectively. These results prove the intuition that using the hybrid model constructed using feed-forward and convolutional layers are more capable of extracting the hidden features and the nonlinear relations between them. Moreover, the results in Fig. \ref{accuracy} outperforms the best results (67.7 \% accuracy) in \cite{BNNLSAP} as shown in table II.
\begin{figure*}[!t]
\includegraphics[width=\textwidth,height=11cm]{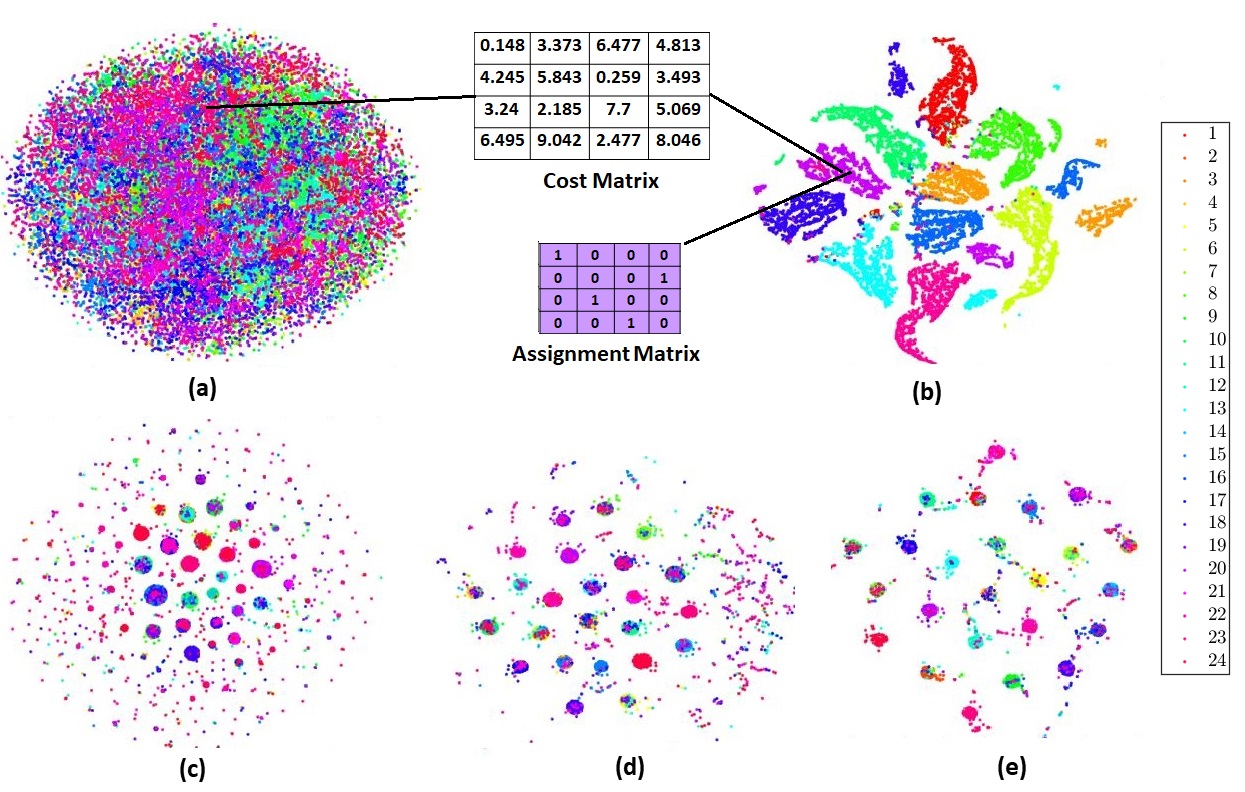}
\caption{Performance of a hybrid autoencoder network of size n = 4 with 24 output clusters.
(a) Visualization of the original 50,000 training dataset samples. Each point represents one pair of inputs (i.e., cost matrix and a label) where different Colors correspond to different labels.
(b) A Sample of 10 clusters and its corresponding cost matrix and label on the map.
(c,d and e) Network clustering performance while training for different epochs 50, 100 and 150 respectively.
}
\label{latent_output}
\end{figure*}


\begin{table}[]
\center
	\begin{tabular}{|l|c|c|c|c|c|l|l|l|}
		\hline
		& \multicolumn{2}{c|}{\begin{tabular}[c]{@{}c@{}}DNN models\\ in \cite{BNNLSAP} acc. (\%)\end{tabular}} & \multicolumn{3}{c|}{\begin{tabular}[c]{@{}c@{}}Proposed models's \\ acc. (\%)\end{tabular}} & \multicolumn{3}{c|}{\cellcolor{Gray}\begin{tabular}[c]{@{}c@{}}Proposed models's\\  time (ms)\end{tabular}} \\ \hline
		n  & FNN                                           & CNN                                           & FNN                          & CNN                          & Hybrid                        & FNN                          & CNN                         & Hybrid                         \\ \hline
		4  & 90.8                                          & 92.76                                         & 92.2                         & 94.5                         & 97.46                         & 3.2                          & 5.5                         & 6.9                            \\ \hline
		8  & 72.3                                          & 77.8                                          & 89                           & 93                           & 94.5                          & 3.6                          & 6.9                         & 7.2                            \\ \hline
		16 & 59.8                                          & 65.7                                          & 83                           & 88.7                         & 92.3                          & 4.2                          & 7.1                         & 7.6                            \\ \hline
	\end{tabular}


\caption{The accuracy and execution time of the three proposed autoencoder models Compared to the classification-based deep network models in \cite{BNNLSAP} for LSAPs with size n =4, 8, and 16. }
\label{Net_results}
\end{table}

Table II shows the accuracy for the proposed autoencoders compared to the classification-based deep neural network model in \cite{BNNLSAP} for different sizes of the problem (i.e., n = 4, 8, and 16). The simulation results in this table show that the accuracy of the three proposed models are better than the classification-based model for all sizes. We notice that the performance of the classification-based model deteriorates with the increase of the size due to the decomposition approach used in \cite{BNNLSAP} which drives the obtained solution fare-away from the optimal solution. On the other hand, the proposed autoencoder deals with the assignment matrix (i.e., the solution) as a whole without any decomposition.

Fig. \ref{loss} shows the convergence of the training loss function of the proposed hybrid autoencoder for a LSAP of size 4 and 8 compared to the best result of the classification-based deep neural network (i.e., for n = 4, batch size = 128, and learning rate = 0.001 ) in \cite{BNNLSAP} as a function of the training epoch. The simulation results show that the proposed hybrid autoencoder converges very fast compared to the classification-based model (150 epochs versus 500 epochs). The simulation results show that the application of the $L_{2}$ loss function and the generative approach is suitable for the assignment problem, especially with the sparsity of the output. the adaptation of the ReLu activation function at the output of the generative model helps the output layer to generate 0 or 1 and converges faster to a less error rates compared with the classification model.


The performance of the proposed autoencoder models can be measured in terms of its ability to cluster (group) different input samples in correspondence to the possible output assignment matrices. The clustering performance of the model can be considered as a measure of the scalability for larger problem size, and its ability to extract features to be able to discriminate between different clusters. Fig. \ref{latent_output} presents the clustering maps for a LSAP of size n = 4, where we have $n ! = 24$ possible solution (i.e., 24 clusters). We track the network discrimination capability during multiple phases (epochs) of the training process to measure the accuracy of assignment matrix generation. A 2D map is generated for each epoch starting with the initial visualization of all the training dataset (50,000 samples in this case) in Fig. \ref{latent_output}.(a). In Fig. Each point represents a sample of an input cost matrix where the color is related to the optimal solutions of that specific sample. A random  cost matrix sample is shown  beside Fig. \ref{latent_output}.(a) to be tracked to the next step in the training process with its associated assignment matrix. The map in (b) shows an intermediate step for only 10 clusters out of the 24 clusters which is followed by maps (c), (d) and (e) at epochs 50, 100 and 150, respectively. It can be noticed that the discrimination power of the network rises while training. Map (e) is taken after the convergence of the network with an accuracy around 97\% where each optimal solution is represented in a cluster and the training sample are grouped to the vicinity of each other. Additionally, the different clusters are well separated and able to be scaled for larger problems.




On the other hand, the proposed architectures have different complexities based on the investigations of the FNN and CNN layers in \cite{he2015convolutional,NovikovPOV15}. Therefore, the proposed FNN-CVAE and CNN-CVAE models have complexities of $O(n)$ and $O(n^2)$, respectively while the Hybrid-CVAE has a complexity of $O(n^2)$ since it is composed of both CNN and FNN layers.
The highlighted part of Table \ref{Net_results} shows the time needed to find the solution averaged over 1000 testing samples for networks with different sizes running on the same machine using the three proposed architectures. The average time ranges from 3.2 ms up to 7.6 ms according to the different depth and number of neurons per layer in each architecture which is much faster than the 0.59 second for the Hungarian algorithm as mentioned in Table I \cite{BNNLSAP} with an accuracy of 97.46\%. Such approach can be used instead of the Hungarian algorithm in different real-time applications.

The computational complexity of the conventional Hungarian algorithm  which is used as benchmark for training the DNN is $O(n^3)$, while the classification-based model  in \cite{BNNLSAP} have a complexity of $O(N^2)$. On the other hand, the proposed architectures have different complexities based on the investigations of the FNN and CNN layers in \cite{he2015convolutional,NovikovPOV15}. Therefore, the proposed FNN-CVAE and CNN-CVAE models have complexities of $O(n)$ and $O(n^2)$, respectively while the Hybrid-CVAE has a complexity of $O(n^2)$ since it is composed of both CNN and FNN layers. Such approaches can be used instead of the Hungarian algorithm in different real-time applications. The total time complexity of all convolutional layers is $O(\sum_{l=1}^{d} n_{l-1} . s_{l}^{2}.n_{l}.m_{l}^{2} )$,  
where $l$ is the index of a convolutional layer, and $d$ is the depth (i.e., number of convolutional layers), $n_l$ is the number of filters (also known as width) in the $l-th$ layer, $s_l$ 	is the spatial size (length) of the filter, and $m_l$ is the spatial size of the output feature map.  Notice this time complexity applies to both training and testing stages, though with a different 	scale, where the training time per cost matrix is roughly three times of the testing time per cost matrix (one for forward and two for backward propagation). 

However, the training phase is usually slow process similar to other machine learning approaches as shown in table \ref{Net_Time}, the training is usually an offline stage, that does not affect the performance enhancement of the proposed solution. Notice that the training is implemented in python 3.6 with Keras 2.1.3 on a 2 Tesla P100 GPU-based machine.

\begin{table}[!h]
	\center
	\begin{tabular}{|l|l|l|l|}
		\hline
		\textbf{n}     & \textbf{FNN} & \textbf{CNN} & \textbf{Hybird} \\ \hline
		4  & 7:35  & 11:20 &         16:11                     \\ \hline
		8  & 12:46 & 19:52 &     22:23                      \\ \hline
		16 & 18:33 & 23:14 &      31:18               \\ \hline
	\end{tabular}
	\caption{The running time for taining networks for LSAPs with size n = 4, 8, and 16. }
	\label{Net_Time}
\end{table}

\section{Conclusion and Future Work}
In this paper, we designed three generative autoencoder models to solve the linear sum assignment problem for resource allocation. A conditional variational autoencoder is training using an optimal dataset generated using the conventional Hungarian algorithm. The main idea is to use a deep learning approach that is capable of mapping the cost matrices of the LSAP to the corresponding optimal solution defined by the assignment matrix. Empirical studies demonstrate the proposed model capability to correctly map the input to the intended output in high accuracy and which is extended to a larger problem sizes. The simulation results show that the proposed hybrid-CVAE approach outperforms the CNN and FNN approaches and the recently introduced deep network models and achieve a high accuracy.

As a future work, we intend to investigate the performance of the proposed model for larger size problems which are challenging due to the large number of possible outcomes. Additionally, we would like to investigate the deep learning approaches to solve other complicated combinatorial optimization problems such as the three-dimensional linear and quadratic assignment problems.

\balance
\bibliographystyle{IEEEtran}
\bibliography{bibliography}

\end{document}